\documentclass[10pt, letterpaper]{article} % DO NOT CHANGE THIS
\usepackage{aaai22}  % DO NOT CHANGE THIS
\nocopyright 
\usepackage{amsfonts}
\usepackage{times}  % DO NOT CHANGE THIS
\usepackage{helvet}  % DO NOT CHANGE THIS
\usepackage{courier}  % DO NOT CHANGE THIS
\usepackage{amsmath}
\usepackage[hyphens]{url}  % DO NOT CHANGE THIS
\usepackage{graphicx} % DO NOT CHANGE THIS
\usepackage[super]{natbib}  % DO NOT CHANGE THIS AND DO NOT ADD ANY OPTIONS TO IT
\usepackage{caption} % DO NOT CHANGE THIS AND DO NOT ADD ANY OPTIONS TO IT
\DeclareCaptionStyle{ruled}{labelfont=normalfont,labelsep=colon,strut=off} % DO NOT CHANGE THIS
\usepackage{float}
\usepackage{algorithm}
\usepackage{algorithmic}
\usepackage[a4paper,
            bindingoffset=0.2in,
            left=1in,
            right=1in,
            top=1in,
            bottom=1in,
            footskip=.25in]{geometry}

\usepackage{newfloat}
\usepackage{listings}
\floatstyle{ruled}
\newfloat{listing}{tb}{lst}{}
\floatname{listing}{Listing}
\title{RECAP-KG: Mining Knowledge Graphs from Raw Primary Care Physician Notes for Remote COVID-19 Assessment in Primary Care}
\author{
    Rachel Lee Mekhtieva$^1$,
    Brandon Forbes,
    Dalal Alrajeh$^1$,
    Brendan Delaney$^2$,
    Alessandra Russo$^1$,
    \\
    $^1$Department of Computing, Imperial College London, London, UK;\\
    $^2$Institute of Global Health Innovation, Imperial College London, London, UK
}
 \affiliations{}
\usepackage{bibentry}
\begin{document}

\maketitle

\begin{abstract}
\noindent \textit{Building Clinical Decision Support Systems, whether from regression models or machine learning requires clinical data either in standard terminology or as text for Natural Language Processing (NLP).  Unfortunately, many clinical notes are written quickly during the consultation and contain many abbreviations, typographical errors, and a lack of grammar and punctuation Processing these highly unstructured clinical notes is an open challenge for NLP that we address in this paper. We present RECAP-KG - a knowledge graph construction framework from primary care clinical notes. Our framework extracts structured knowledge graphs from the clinical record by utilising the SNOMED-CT ontology both the entire finding hierarchy and a COVID-relevant curated subset. We apply our framework to consultation notes in the UK COVID-19 Clinical Assessment Service (CCAS) dataset and provide a quantitative evaluation of our framework demonstrating that our approach has better accuracy than traditional NLP methods when answering questions about patients. }
\end{abstract}

\section{Introduction}

\noindent Clinical decision support systems for differential diagnosis are designed to aid the process of diagnosing patients. Primary care physicians work with an increasingly large number of patients during a restricted time frame, after which they must assign an appropriate diagnosis to the patient and outline the next treatment steps. Clinical decision support systems could facilitate this process, thus helping improve patient outcomes and reduce the burden on the clinician. However, traditional regression-based decision support systems mostly rely on structured data (patient age, sex, smoker status, etc.) and do not use unstructured consultation notes as input. The complexity of processing the notes is the main factor that leads to their information potential being entirely omitted in clinical decision support systems. Therefore, these systems potentially miss nuance and risk delaying appropriate care and misdiagnosing the patient. Other differential diagnosis generators have been built on machine learning methods but provide no explanation for their diagnosis \cite{Dai202021DeepPatients} \cite{Liu202020ADiseases}. 

Structured knowledge graph extraction from medical notes can provide a practical use for harnessing dense information about patients and diagnoses. This information is currently omitted from all clinical decision support systems due to the complexity of processing medical notes and the fact that access to the notes is usually completely restricted, as free text medical notes may contain personal identifiers for the patient or third parties and cannot be reliably de-identified.

The area of graph construction from unstructured text data has been widely explored\cite{Melnyk2022KnowledgeText} \cite{Friedman2022FromApproach} \cite{Yu2020AutoKG:Answering}. Domain-specific methods have been developed for a multitude of applications: cybersecurity\cite{Agrawal2022BuildingEducation} \cite{Jia2018ACybersecurity} manufacturing\cite{Yan2020KnowIME:Equipment}, cultural heritage\cite{inbook}, etc. The more common methodology is to apply natural language processing to input data in order to perform entity recognition and relation extraction, then construct a knowledge graph from these outputs. However, such a pipeline is not easily applicable to medical notes due to several factors specific to this application domain. Firstly, the consultation notes are written in haste, especially in the case of phone consultations, and therefore they often do not follow grammatical rules and proper sentence structure in the English language. Passages missing the subject, object, and connecting words are common. The natural language includes specialist medical vocabulary that normally would not be easily recognised by conventionally trained entity recognition models. 

In this paper we propose a novel framework that is able to perform knowledge graph extraction from raw primary care physician notes. Our method - RECAP-KG, uses supported facts and sentence parse trees in order to generate knowledge graphs from consultation notes that were written after consulting COVID-19 patients as part of the UK pandemic response COVID-19 Clinical Assessment Service. This was a telephone advice and triage service for patients contacting the ambulance and emergency services with COVID-19 symptoms. We show that our framework is able to process a subset of consultation notes of COVID patients, extracting relevant disease information, including signs/symptoms. Specifically:

\begin{itemize}
    \item Our framework constructs passage-level knowledge graphs from medical consultation notes by relying on sentence parse trees and support phrases for diseases, signs, symptoms, and other relevant information. The support phrases are mined from the SNOMED\cite{El-Sappagh2018SNOMEDScience} ontology, as well as adapted from values used in the RECAP\cite{Espinosa-Gonzalez202022RemoteStudiesb} (REmote COVID-19 Assessment in Primary Care) patient risk prediction tool.
    \item We evaluate our model in two ways: (i) firstly, in order to demonstrate that our knowledge graph extraction mechanism preserves the semantics of the original sentence, we validate knowledge graphs by translating them back into natural language and comparing the generated sentences to the original pre-processed sentences, achieving mean cosine similarity of 73\% (ii) we then evaluate the performance of the whole pipeline with respect to well-known NLP models, such as BERT, by facilitating transparent fully interpretable Question Answering tasks over the learned knowledge\cite{Zhang2019BERTFQ}.
\end{itemize}

\section{Related Work}
% Knowledge graph construction from unstructured text is a well-explored problem. In addition to domain-specific knowledge graph construction methods (\cite{Melnyk2022KnowledgeText}, \cite{Friedman2022FromApproach}, \cite{Jia2018ACybersecurity} etc.), generic (i.e. domain-independent) knowledge graphs have been steadily evolving and widely used for domain-independent tasks. Examples of generic knowledge graphs include Google KG\cite{singhal2012introducing}, BabelNet\cite{}, and DBPedia\cite{}.

 Knowledge graphs constitute a powerful tool in knowledge representation due to their underlying structure that facilitates better reasoning, comprehension, and interpretation\cite{DBLP:journals/corr/abs-2011-00235}. They are therefore used for a variety of domain-specific and generic tasks. Domain-independent knowledge graphs have been steadily evolving and are widely applied to solving generic tasks at present. Examples of such domain-agnostic knowledge graphs include large open-knowledge graphs, such as Google KG \cite{singhal2012introducing}, Wikidata \cite{Vrandecic2014Wikidata:Knowledgebase}, BabelNet \cite{Navigli2012BabelNet:Network}, and DBPedia \cite{Exner2012EntityEF}. Open-knowledge graphs often use collaborative crowdsourcing (e.g. Wikidata and DBPedia). However, these large-scale graphs are by and large unsuitable for specific narrow-domain tasks, since they often won't have enough nuance to represent domain-specific and/or fine-grained relations. Therefore, constructing domain-specific KGs remains an essential task since these graphs have semantically intertwined applications with problems in their respective domains\cite{DBLP:journals/corr/abs-2011-00235}. Furthermore, domain-specific knowledge graphs often incorporate human expert knowledge that is crucial for solving domain-specific problems \cite{Kejriwal2019Expert-guidedRules}. On the other hand, Wikipedia and information on the Web, which are used as information sources for generic KGs, would inevitably lack such specific insight. Both domain-specific and domain-agnostic knowledge graph construction methods may utilize unstructured text as a source of relevant information. The ability to extract structured knowledge from text is a desirable property for various tasks (e.g. search \cite{singhal2012introducing}, education \cite{Agrawal2022BuildingEducation}, etc.). 
 
 Knowledge graph construction from unstructured text is a well-researched problem. The vast majority of KG construction methods are domain-specific and use some knowledge base. Attempts have been made to create domain-agnostic knowledge graph generation models, such as the end-to-end neural model for domain-agnostic knowledge graph construction from text \cite{Stewart2020Seq2KG:Text}. However, a domain-agnostic method that would consistently outperform its domain-specific counterparts is yet to be seen. Stewart et al. note that conventional domain-specific KG construction methods still achieve the best overall performance when compared to the end-to-end domain-agnostic model and therefore remains state-of-the-art. Furthermore, domain-specific KG construction methods typically rely on an ontology curated for the specific application domain.
 
 The healthcare sector is among the specialist domains that require expert involvement and application-specific knowledge when consolidating meaningful information in a structured format. Moreover, knowledge graphs provide a convenient data representation model for large volumes of heterogeneous medical data\cite{Zhang202020HKGB:Incorporated}. Medical ontologies, such as SNOMED CT \cite{El-Sappagh2018SNOMEDScience}, are used to guide the KG construction process. At present, natural language processing (NLP) is the most widely used big data analysis method in healthcare \cite{Mehta2018ConcurrenceReview}. However, the vast majority of knowledge graph construction methods in healthcare gather information from structured data in electronic health records (EHR). The methods that process unstructured health data, such as the KG curation framework for subarachnoind hemorrage by Malik et. al. \cite{Malik2020AutomatedPhenotype}, typically include progress notes, discharge summaries, and radiology reports. Harnoune et. al. use the Bidirectional Encoder Representations from Transformers (BERT) model for knowledge extraction from clinical notes \cite{Harnoune2021BERTAnalysis}. To date, there are no knowledge graph construction frameworks in healthcare that process primary care physician consultation notes - the first point of contact for a patient seeking non-urgent medical help. Holmgren et. al. note that US clinicians receive more system-generated messages and write a higher proportion of automatically generated note text consultation notes compared to their UK counterparts\cite{Holmgren2021AssessmentSystems}. Therefore, in the context of UK consultation notes that are relevant to our work, a distinction can be made between progress notes and consultation notes, with the former being more unstructured due to the rushed nature of primary consultations.
\section{Method}.

\begin{figure}[h]
    \centering
    \includegraphics[width=0.9\columnwidth]{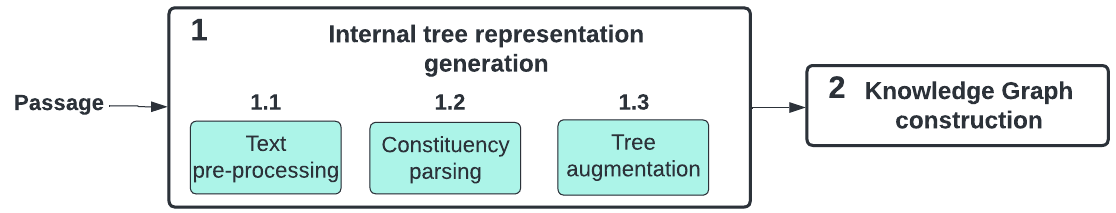}
    \caption{We begin with the text from primary care physician consultation notes. In the first stage of the pipeline, we generate an internal tree representation for the passage in three main steps: (1.1) the text is split into sentences and pre-processed. (1.2) Then, a constituency parse tree is generated for every pre-processed sentence in the text. (1.3) Finally, node replacement and information-hiding techniques are applied to the constituency parse trees in order to complete the internal tree representation. In the second stage, a knowledge graph is extracted from the resulting tree representation (2).}
    \label{fig:pipeline}
\end{figure}

We outline the RECAP-KG framework. The knowledge graph generation pipeline (Figure \ref{fig:pipeline}) consists of two main stages: (1) internal tree representation generation and (2) knowledge graph construction. 

% We outline the RECAP-KG framework. The knowledge graph generation pipeline (Figure \ref{fig:pipeline}) consists of four main stages: (1) sentence pre-processing, (2) constituency parsing, (3) parse tree augmentation, and (4) knowledge graph construction. Our method uses \textit{supported facts} - a dictionary of relevant medical terms. These terms are extracted from a medical ontology prior to running the main pipeline and are used for graph construction, adding an extra step (0) to the framework. The supported facts only need to be generated once.

 \subsection{Internal Tree Representation Generation}
 We propose an \textit{internal tree representation} for our method. Using trees as the core structure in our model has proven to be the most versatile approach for data representation for the given task, since a tree (1) is the natural output of many standard NLP tasks, such as constituency parsing, (2) can be extended to token-level tasks by adding token annotations to the leaves of the tree, as in constituency parsing the leaves of the tree represent word tokens, (3) is the most straightforward data representation to transform into a knowledge graph.

\begin{figure}[h]
    \centering
    \includegraphics[width=0.9\linewidth]{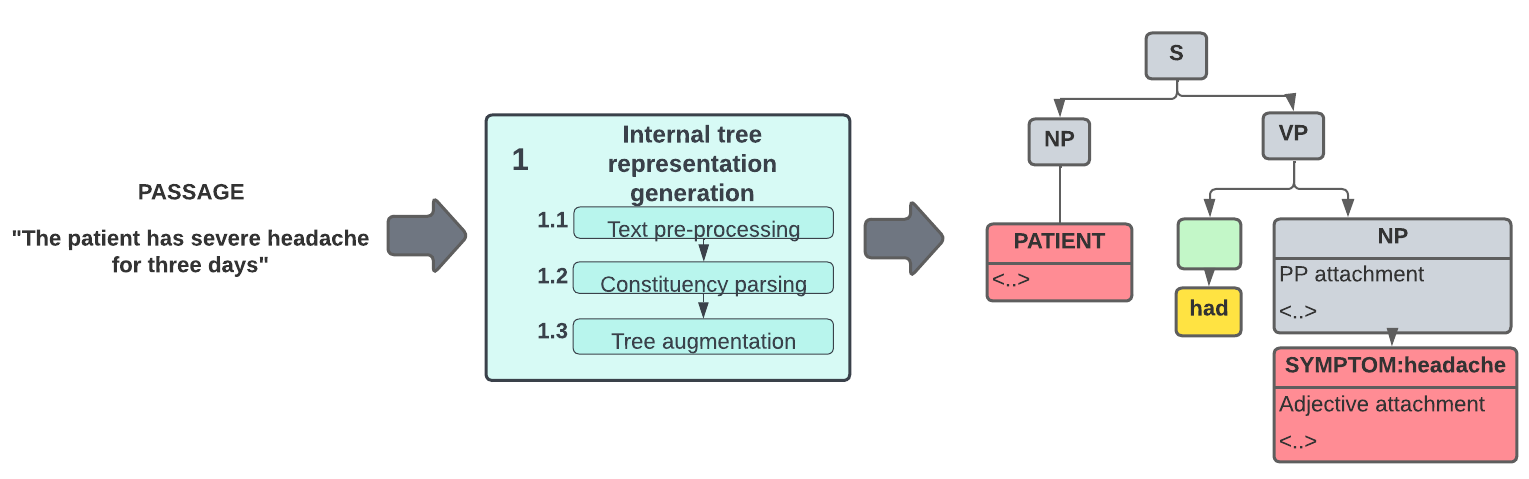}
    \caption{An example of the internal tree representation built for the sentence "The patient had severe headache for three days".}
    \label{fig:augmentation_prev}
\end{figure}

 Our internal tree representation is a modified version of a \textit{constituency parse tree}. The internal tree representation always contains a \textit{noun phrase node} with a single patient entity node and a \textit{verb phrase node} that contains the verb "has" and a noun phrase with one or more symptom entity nodes. The \textit{symptom entity node(s)} may have zero or more \textit{attachments} - sentence constituents that add additional information to the symptoms, such as their severity. The tree may also have \textit{list nodes} that encapsulate conjunctive or disjunctive lists of symptom nodes or their attributes. Finally, the internal tree representation may have independent phrase and word token nodes that cannot be attached to any symptom and preserve the original semantics of the sentence. Figure \ref{fig:augmentation_prev} shows an internal tree representation example.

\paragraph{Pre-processing and Constituency Parsing}\label{parse_tree_generation}
We now outline the pre-processing necessary to translate the text data into our internal tree representation. 

To begin with, we use a \textit{constituency parser} to generate the base for our internal tree representation. A constituency parser breaks down a sentence into sub-phrases, known as constituents. The root of the tree represents the full sentence. The leaf nodes always contain the original word tokens from the sentence. Each word's direct parent node contains its \textit{part-of-speech tag} (noun, verb, adverb, etc.). All the other nodes represent the \textit{constituents} of the sentence (e.g. noun phrase). The fact that a constituency parse tree contains both the word tokens and their part-of-speech tags, as well as constituents that reflect the structure of the original sentence, makes it a convenient tree representation for knowledge graph construction. In our implementation, we use the AllenNLP constituency parser \cite{Gardner202018AllenNLP:Platform}. 

Before applying a constituency parser, we pre-process our input text (Figure \ref{fig:pipeline} Step 1.1). This pre-processing is needed since (i) constituency parsing is only defined at the sentence level, whereas a physician note may contain multiple sentences, (ii) most sentences in the original note are incomplete (e.g. missing the verb etc.), (iii) typos and abbreviations are prevalent in primary care consultation notes and need to be corrected/expanded for better performance of the constituency parser.

We perform pre-processing in two main steps: (1) spellcheck and (2) passage splitting. The first stage is straightforward: typos and spelling mistakes are corrected in the passage, and shorthand notation and conventional medical abbreviations are expanded to their respective full form. In the second stage, the notes are split into sentences. 

Since medical summaries are written in brief notation that often does not have a structure of proper written English, after the passages are split into shorter segments at every delimiter, each segment is expanded into a sentence that contains at least a subject and verb. In order to complete a sentence, we first generate a constituency parse tree for every segment. At present, the sentence expansion is done via simple pattern-matching rules to insert the subject into the sentence, if missing. Pattern matching is then performed on these parse trees. In total, we make use of four patterns: a negated noun phrase ("no fever"), a noun phrase or list of noun phrases ("fever, fatigue, and anxiety"), an adjective phrase ("anxious"), and verb phrases ("coughing", "can talk" etc.). If a pattern is matched, we add the subject. For example, a negated noun phrase "no breathlessness" would be expanded to "The patient does not have breathlessness".  

In this work, we restrict knowledge graph construction to sentences that can be expanded to the form 'The patient has.." (e.g. "The patient has fatigue and severe nausea") or "The patient is.." (e.g. "The patient is breathless"). For our application, perfect grammatical correctness (i.e. "The patient has had headache for two days", as opposed to "The patient has headache for two days") is not required, as the constituency parse trees are only generated as a basis for our internal tree representation used during knowledge graph extraction.

Finally, a constituency parse tree is re-generated for each expanded sentence (Figure \ref{fig:pipeline} Step 1.2). The final output of the pre-processing stage for each text passage is a list of constituency parse trees of complete, grammatically correct sentences. 

\subsubsection{Parse Tree Augmentation}

\begin{figure}[h]
    \centering
    \includegraphics[width=0.9\linewidth]{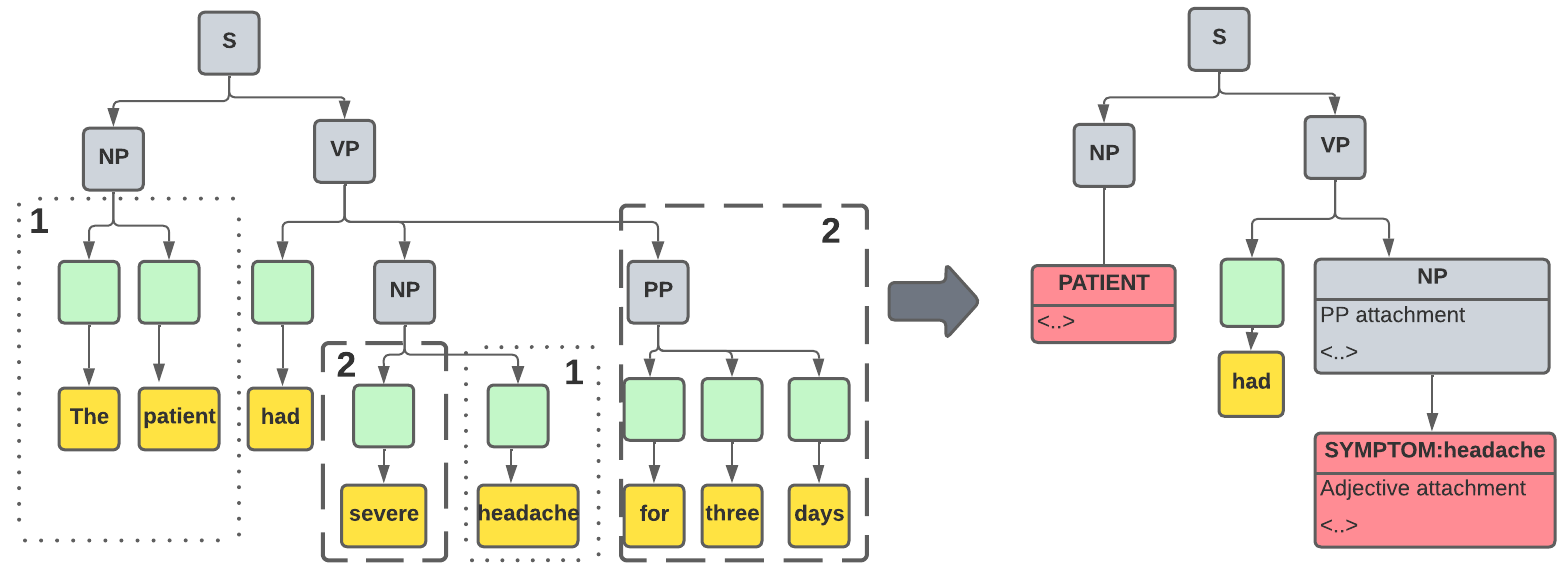}
    \caption{A tree augmentation example of a constituency parse tree built for the sentence "The patient had severe headache for three days". (1) First, special entity nodes are generated for the patient and the symptom and replace their respective word token components. The original tokens and their pos tags are maintained in node attributes (2) Secondly, attachments that add additional information to the main sentence are detected and hidden in node attributes. Note that the POS tags and the full list of node attributes are omitted in the diagram for simplicity. The original parse tree structure is retained in node attributes.}
    \label{fig:augmentation}
\end{figure}

Our knowledge graph extraction algorithm relies on accurate internal tree representation as input. To generate these trees, we use the outputs of multiple models to encode semantic information into our internal representation. Since the constituency parser output alone does not accurately represent several properties required to accurately encode the information contained in the original text into a knowledge graph, such as information hiding and symptom entity nodes, we define additional pre-processing steps for parse tree augmentation (Figure \ref{fig:pipeline} Step 1.3). The goal is to apply a series of processing stages to convert constituency parse trees generated by AllenNLP into a list of well-structured internal tree representations to be used for knowledge graph construction. An example of tree augmentation is presented in (Figure \ref{fig:augmentation}).

\paragraph{Node Generation}

Node generation involves modifying our internal tree representation by inserting additional nodes that were not originally in the constituency parse output and/or replacing one or more nodes with a replacement node. Note that these are nodes in the internal tree representation, not the final knowledge graph. The generated nodes capture specific semantic patterns in the constituency parse tree. We, therefore, refer to them as \textit{semantic nodes}. The motivation behind semantic nodes is the ease of interpretation for the knowledge graph extraction algorithm, as they encode specific relations that the final knowledge graph should represent.

% Therefore, these new nodes can be further grouped into (1) Leaf Nodes and (2) Semantic Nodes according to their purpose. 

    % \item \textbf{Leaf Nodes} are nodes that represent atomic concepts i.e. compound phrases that semantically should not be split up any further. The motivation behind leaf nodes is to refine the output of the spaCy tokeniser used by AllenNLP that we use to generate the constituency parse. Specifically, spaCy would split phrases, such as "on-call", into two tokens. However, in the knowledge graph, we would like to interpret "on-call" as a single noun-phrase. Similarly, phrases such as "Magnetic Resonance Angiography", as processed by the tokenizer, would contain several relations and would therefore not be treated as a single entity by the knowledge graph extractor, but a set of entities and relations. Should the graph be queried for the entity "Magnetic Resonance Angiography", no entity nodes would be identified. Therefore, we would generate a single leaf node for such phrases, replacing the multiple word tokens generated by the tokeniser. Note that the leaf node still retains the replaced tokens as node attributes, should they be required in the later stages of knowledge graph generation.

Semantic nodes are divided into 4 types: (i) \textbf{Patient entity node} - the single subject, (ii) \textbf{sign/symptom entity nodes} that represent spans in the text that match a supported fact and thus represent a sign/symptom/pre-existing condition, and (iii) \textbf{list nodes} that describe a conjunctive or disjunctive list, such as "fatigue, nausea, and headache". List nodes are especially useful since they simplify our representation through information hiding.
   
\subsubsection{Attachments}

The second step of parse tree augmentation handles \textit{attachments}. Attachments refer to constituents in the sentence that do not directly define the main action/fact of the sentence but add additional information to the main action or to other attachments. Therefore, these constituents do not need to be parsed at the same time as the main action. This property is useful since many pre-trained models, including the AllenNLP constituency parser, demonstrate significantly better performance on smaller inputs, as opposed to larger ones. Therefore, processing constituents of the sentence separately allows us to improve the overall performance of the pipeline. We perform a top-down traversal of the parse tree, identifying attachments (e.g. prepositions, adjectives, adverbs). When identified, we temporarily hide the attachments from the tree and instead store them as attributes of the core sentence structure. Therefore, from the perspective of the constituency parser, the attachment has been removed from the parent node; however, it is still present in the internal tree representation, so the accuracy improves at no cost to information completeness. 

\subsection{Knowledge Graph Construction}
\begin{figure}[h]
    \centering
    \includegraphics[width=0.9\columnwidth]{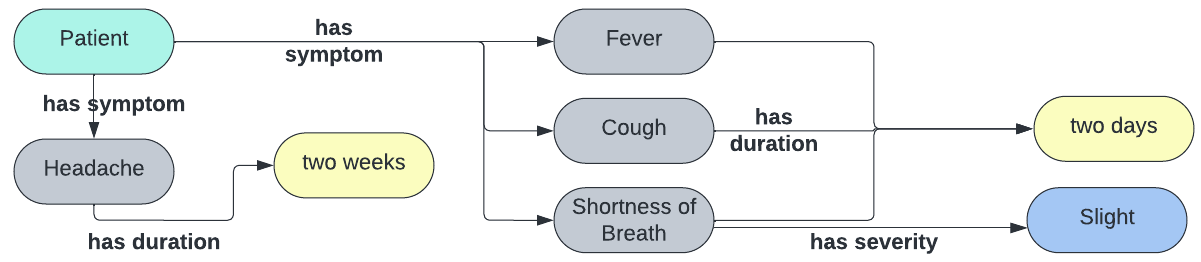}
    \caption{A knowledge graph built by RECAP-KG for the consultation note: "headache couple weeks last 2 days more feverish and cough slight sob, needs to take deep breaths, even on minimal effort."}
    \label{fig:graph}
\end{figure}

The knowledge graph construction stage of our implementation uses as input a list of augmented parse trees. The output is a respective knowledge graph per each consultation summary \textbf{passage} (Figure \ref{fig:graph}). We rely on a knowledge base of \textit{supported facts} for knowledge graph construction. 

\paragraph{Supported Facts Generation}
 Our method uses \textit{supported facts} - a dictionary of relevant medical terms. These terms are extracted from a medical ontology prior to running the main pipeline and are used for graph construction. The supported facts only need to be generated once. They are required in order to focus on the specialist vocabulary relevant to the application at hand by generating more compact knowledge graphs. Supported facts include expert-curated signs and symptoms, as well as medical vocabulary mined from the SNOMED CT ontology \cite{El-Sappagh2018SNOMEDScience}. Specifically, the motivation for including supported facts in our approach is threefold: (1) to ensure that highly specific medical vocabulary (e.g. "cholangiopancreatography") is recognised, independent of the underlying NLP method used to parse the original text and (2) to identify medical terms that should be interpreted by the knowledge graph as atomic concepts, as opposed to sets of words with relations (e.g. "magnetic resonance cholangiopancreatography"). For each relevant medical concept (supported fact), we define the main label - the text that would appear on the corresponding node in the knowledge graph, noun and adjective supports - terms that are synonymous with the concept. 
 
 The supported facts in our method are generated in two ways: 
 \begin{enumerate}
     \item Signs and symptoms that may potentially be important are mined from the SNOMED-CT ontology\cite{El-Sappagh2018SNOMEDScience} through BioPortal\cite{Whetzel202011BioPortal:Applications}. SNOMED-CT is a polyhierarchical terminology system that is in wide use worldwide, including the NHS. SNOMED-CT provides a robust and complete set of clinical terms that may be of interest in any given consultation summary. Although the ontology is highly expressive, clinical records may have both terminology and free text present with a sparse and quite arbitrary selection of specific terms used by clinicians. Therefore, natural language processing is still required as opposed to simply matching ontology terms.

    \item Signs and symptoms that are considered to be highly relevant to the clinical presentation may also be provided by the clinician for a given medical setting. Note that this part is not essential for the approach but may be useful in practice, as for most conditions the main signs and symptoms are known. In our case, we relied upon the RECAP research programme, which had derived a set of concepts relevant to acute COVID-19 assessment \cite{Espinosa-Gonzalez202022RemoteStudiesb}.

 \end{enumerate}

\paragraph{Knowledge Graph Structure}
We now present the knowledge graph extraction phase of the pipeline (Figure \ref{fig:pipeline} Stage 4), which extracts the knowledge graph from our internal Tree representation of the original passage. First, we propose a knowledge graph structure for primary care physician notes. The nodes in the graph represent entities, and the edges represent relations between said entities. The information that we need to capture in the knowledge graph about a patient is what symptoms/signs the patient has, how severe these symptoms are, and for how long the patient has had them. The graph, therefore, contains four types of nodes: (I) the \textit{patient entity node} that corresponds to the subject of the medical summary, (ii) \textit{symptom/sign} entity nodes that represent any clinical condition/symptom attributed to the patient in the summary (e.g. shortness of breath, nausea, fever), (iii) \textit{symptom duration} nodes that denote how long a corresponding symptom has been present (e.g. 'a week', 'two days ago'), and (iv) \textit{symptom severity} nodes that encode the severity and other properties of a corresponding symptom (e.g. 'slight', 'severe', 'intermittent', 'continuous'). 

At present, RECAP-KG input vocabulary is defined as sentences that begin with 'the patient has', as well as phrases that can be extended to this form. An example of a graph built by RECAP-KG is shown in Figure \ref{fig:graph}.

\subsubsection{Knowledge Graph Extraction}

We now describe the process of extracting a knowledge graph from our internal parse tree representation. The knowledge graph extraction algorithm takes as input a list of augmented parse trees for the original text and individually extracts entity and relation instances from each tree. In order to do this, the algorithm heavily relies on \textit{open information extraction (OpenIE)} - the process of extracting relations from text. Most importantly, OpenIE does not rely on a pre-defined schema in order to extract relations between two instances and can be applied to any kind of text. Therefore, it is applicable to both the complete passage and its smaller constituents, such as the core sentence and its attachments. Thus, instead of extracting OpenIE relations from the whole original constituency parse tree, we will work with smaller segments, constructing sub-graphs. In our implementation, we use the AllenNLP OpenIE library\cite{Gardner202018AllenNLP:Platform}. 

Firstly, we extract the main sub-graph for the core sentence, i.e. the sentence excluding attachments. We then extract sub-graphs for each attachment phrase and re-connect them with the main sub-graph. Finally, we remove nuisance relations as OpenIE frequently returns argument-less relations, specifically for various forms of the verb 'to be'. This completes the graph extraction stage for a single sentence in a passage. This process is then repeated for every parse tree corresponding to sentences in the original text, and any new instances and relations not yet represented in the graph are added. The nodes in the graph do not repeat, even if they are mentioned in multiple sentences. For example, there can only ever be one 'patient' node for every text. Thus the extracted instances for each parse tree are then combined into a single graph representing the whole text. 

The final output of the pipeline is a knowledge graph where the nodes are either entity nodes representing the patient and support phrases (signs/symptoms) or regular nodes corresponding to relevant words of the sentence parse tree. The edges record semantic relations between the nodes, such as a patient having a symptom etc.

\section{Evaluation}

We want two answer two questions: (1) how well are the semantics of the original primary care physician note preserved in the knowledge graph? and (2) is our bespoke KG extraction method as good as a language model for text understanding? We, therefore, evaluate the performance of our knowledge graph construction pipeline using two methods for the first and second questions respectively: 

\begin{enumerate}
    \item \textbf{Sentence re-construction}: we translate generated knowledge graphs into text and evaluate preserved semantics of the original passage.
    \item \textbf{Consultation note comprehension}: we translate the knowledge graphs into a symbolic representation, along with a set of predefined questions that a clinician might want to ask about a patient. We then run a symbolic solver on the result and compare the correctness of the answers with the performance of Bidirectional Encoder Representations from Transformers (BERT) pre-trained for question answering \cite{Zhang2019BERTFQ}. This method allows us to assess the text understanding of RECAP-KG by comparing it to one of the most widely used NLP models.
\end{enumerate}

\subsection{Dataset}
We evaluate our model on CCAS - a dataset that contains structured and unstructured retrospective data on 315 000 COVID patients in the UK. The dataset includes primary care physician consultation notes, as well as patient outcome (recovery, hospitalisation, death etc.). We use a subset of CCAS with 2004 entries for the evaluation which includes passages that contain at least one sentence that can be extended to the form "Patient has ..'. Access to the CCAS data was obtained under approval from the Health Research Authority under section 251 of the NHS Act 2006. Privacy protection of the data was maintained by extracting text records from the CCAS EHR system, data then being linked per patient using NHS number to health data on hospital admissions and deaths (Hospital Episode Statistics) by NHS Digital. A hashing algorithm was used by Oxford university to Pseudonymise the data which was then made available for analysis in the ORCHID Secure Environment. Prior to making the data available, the Python routine Scrubadub \cite{scrubadub} was used to largely remove personal identifiers. HRA accepted that the remaining risk of disclosure was low and outweighed by the public benefit of using the data in the context of the COVID-19 pandemic.

\subsection{Sentence Reconstruction}
In this section we describe the process of converting a generated knowledge graph back into text with the goal of comparing this text with the original input passage and thus evaluating whether sentence semantics are preserved. This step takes as input a knowledge graph generated by RECAP-KG. It then builds a sentence in natural language attempting to re-create the original pre-processed sentence from which the graph was built. The decoder is a simple json parser that iterates through the nodes of the graph and translates them according to the following rules:

\begin{itemize}
    \item The “core” nodes of the graph are: (a) the patient node and (b) the sign/symptoms node. First, these nodes are located and translated into their equivalent phrases. Then, the relation between the two nodes ('has symptom' or 'does not have symptom') is inserted between the two phrases.

    \item The “descriptor” nodes add characteristics to the recognised symptoms (“severe shortness of breath”, “wheeze when talking” etc.). In essence, these nodes represent the entities originally contained in attachment segments. Similarly to the core nodes, the start and end of the relation are translated into natural language with their respective relation inserted between the two. 
\end{itemize}

 We evaluate how close the reconstructed passage is to the original through \textit{semantic text similarity score (STS)}. The score is calculated by computing the cosine similarity between BERT\cite{Devlin2018BERT:Understanding} embeddings for the original and reconstructed sentences. The STS scores are then averaged as follows: $\frac{1}{n}\sum_{i=1}^{n}STS(s_1, s_2)$, where $n$ is the number of passages in the original dataset.

\subsubsection{Results} At present, on 2004 examples we achieve a 79\% mean STS score when comparing passages reconstructed from knowledge graphs and the original passages.

The loss of accuracy is due to two reasons. Firstly, the original sentence inevitably contains more tokens than can be inferred from the knowledge graph. These tokens sometimes include nuisance information, such as 'wife called 111', that would by design be ignored by the knowledge graph construction algorithm. However, from the perspective of semantic similarity between the original passage and the newly constructed sentence from the knowledge graph, this would result in a loss of accuracy. This can be addressed by designing a better heuristic for comparing the two sentences, such as assigning lower significance to constituents of the original sentence that contain nuisance information. Secondly, at present, the set of sentences from which our method can construct knowledge graphs is restricted to sentences of the form 'The patient has ..'. However, we include all passages that contain \textit{at least one} such sentence when evaluating the performance. The rest of the passage may include sentences of a different structure and knowledge graphs would not be generated for these sentences. Therefore, when translating the knowledge graph back into text, the semantic information originally contained in those sentences will be lost. This can be addressed by expanding our approach to other sentence structures ("Patient will ..", "Patient's X is Y", etc.) or generalising it to be sentence structure agnostic.

\subsection{Consultation Notes Comprehension}
In addition to evaluating preserved semantics of the original passage, we wish to evaluate whether our bespoke knowledge graph construction method is as expressive as a language model. Unlike our method, language models are not interpretable. We aim to assess the difference in performance between the two. We now describe the question-answering evaluation method where our model will be compared to BERT\cite{Devlin2018BERT:Understanding} for question answering, trained on SQuAD\cite{Rajpurkar2016SQuAD:Text} - a dataset containing over 100 000 questions for machine comprehension of text. 

In order to perform the evaluation, we have produced the questions that are related to the text that we would like our system to understand. The questions include eight yes/no questions and two multiclass questions that a healthcare professional might want to ask about a COVID-19 patient (Table \ref{tab:questions}).

\begin{table*}[h]
    \centering
    \begin{tabular}[width=0.6\columnwidth]{| c c c|} 
     \hline
     1 & Question & Type \\ [0.5ex] 
     \hline\hline
     2 & Does the patient have shortness of breath? & Y/N  \\ 
     \hline
     3 & Does the patient have a fever? & Y/N   \\
     \hline
     4 & Is the patient breathless? &  Y/N  \\
     \hline
     5 & Does the patient have fatigue? &  Y/N  \\
     \hline
     6 & Does the patient have a rash? &  Y/N  \\ 
    \hline
     7 & Does the patient have a headache? &  Y/N \\
     \hline
     8 & Does the patient have a wheeze? & Y/N  \\
     \hline
     9 & Does the patient have confusion? & Y/N  \\
      \hline
     10 & How severe are the symptoms? &  Multiclass \\
      \hline
     11 & When did the symptoms first start? & Multiclass\\ [1ex] 
     \hline
    \end{tabular}
    \caption{The questions used for the question answering part of our evaluation.}
    \label{tab:questions}
\end{table*}

\begin{figure}[]
    \centering
    \includegraphics[width=\columnwidth]{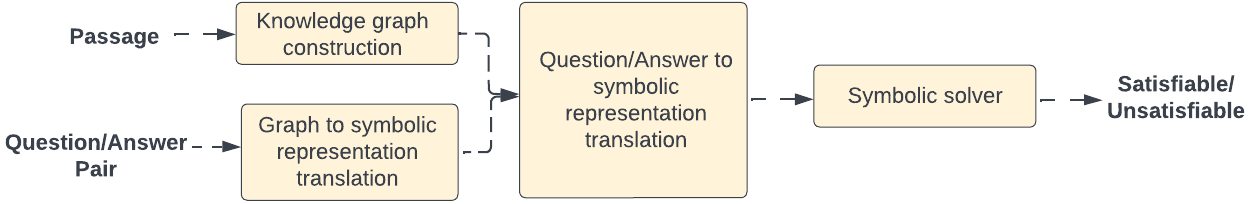}
    \caption{First, we generate a knowledge graph for each entry in the data (1). Then, we translate the knowledge graph (2) and a question-answer pair (3) into their respective symbolic representations. Finally, we combine the two translations and run the result through a symbolic solver to generate a result. The result will either show 'satisfiable', in which case the information of the graph matches the ground truth, or 'unsatisfiable', in which case the information represented in the graph is incorrect and/or insufficient to answer the question correctly, as opposed to the original passage}
    \label{fig:pipeline_qa}
\end{figure}

In order to perform question answering on our model, we translate the knowledge graphs, as well as question-answer ground truth pairs into a symbolic representation. The question-answer pairs are prompts for the language that has been extracted from the text. The formal definitions are omitted from this paper due to their complexity. The translation produces a symbolic program where the question is encoded in zero or more background rules, the ground truth answer is encoded in a constraint, meaning that the program will fail if the ground truth is contradicted. The knowledge graph is encoded as a set of literals (symbolic facts). The program is then executed in the Answer Set Programming solver \cite{ASP}, which produces a result that can either be "satisfiable" - meaning that the facts extracted from the knowledge graph are consistent with the ground truth or "unsatisfiable" - meaning that there is a contradiction between the extracted facts and the ground truth constraint meaning that the information encoded in the knowledge graph is therefore incorrect (Figure \ref{fig:pipeline_qa}).

\subsubsection{Results}
We evaluate the performance of the two models on two sets of yes/no and multiclass questions. The questions are grouped according to the answer type - in theory, the models should perform similarly on every question in a given group. The evaluation metric for the yes/no questions is accuracy, defined as

$$ \frac{TP+TN}{TP+FP+FN+TN}$$

where TP/TN stands for 'true positive/negative' respectively and FP/FN stands for 'false positive/negative' respectively. 

For multiclass questions, the metric used is macro averaged precision, where precision is calculated for all classes individually and then averaged, defined as:

$$ \frac{1}{n}(\frac{TP_1+TP_2+..+TP_n}{TP_1+TP_2+..+TP_n+FP_1+FP_2+..+FP_n})$$

where a true positive denotes an example attributed to a class and indeed being of that class and n is the number of classes. Macro averaged precision is selected since it takes class imbalance into account, which is prevalent in health data (e.g. patients who had symptoms for longer than a year as opposed to for two weeks).

We note from the results (Table \ref{tab:qa_results}) that our bespoke KG-construction method RECAP-KG consistently outperforms BERT for question answering. However, we do not achieve perfect accuracy, despite using a restricted subset of the data that our method has been curated for. This is due to the fact that our knowledge graph construction algorithm relies on good-quality parse trees, which are in turn reliant on good sentence structure. We enforce this sentence structure by splitting the original passage into chunks and thereafter extending each chunk to a full sentence by adding the subject and the verb if the verb is missing too (Figure \ref{fig:augmentation_prev} Steps 1.1, 1.2). However, our current sentence extension method is limited - it looks for a finite set of patterns in the constituency parse tree for the incomplete sentence only up to the first child of the root, under the assumption that the passage has been split into short chunks. When the assumption holds, the algorithm would find a match for one of the patterns, which is indeed the case for most entries in the data, as demonstrated by the high accuracy in the results. However, if not found, it may fail to add the subject or default to an incorrect verb form (e.g. 'patient is' instead of 'patient has'). This is especially relevant for notes in primary care. Entries, where information about a patient is written in one long sentence with no conjunction/disjunction, are prevalent. We address this by splitting the passages at every delimiter. However, if no delimiters are present either, the passage containing one long sentence would not be split into chunks, passing a complex constituency parse tree to the sentence extension stage. The sentence extension stage might then fail to extend the sentence correctly, which would result in an incorrect internal tree representation. 

An example of such an error case from the CCAS dataset is: "no cough tickle only at start first few days sweaty 36 never over 37 headaches diarrhoea not sleeping much not eating drinking breathlessness back and chest feel uncomfortable unable to take deep breath struggling to complete sentence lying in bed mostly struggles to get up stairs and exhausts here no phlegm toilet on same floor as bedroom and able to get there herself". Here, the subject is missing, but the constituency parse tree generated from the chunk would be too complex to match onto one of the pre-defined patterns. This can be improved by creating a better sentence extension method that would be able to perform pattern-matching recursively on the constituency parse tree and select the best pattern, rather than stopping at the first noun/verb phrase. 

\begin{table}[h]
    \centering
    \begin{tabular}[width=0.8\columnwidth]{|c|c c|}
     \hline
     Model & Yes/No & Multiclass \\ [0.5ex] 
     \hline\hline
     BERT & 53\% & 49\%\\ 
     \hline
     RECAP-KG & 95.2\% & 88.2\%\\
     \hline
    \end{tabular}
    \caption{Comparison of RECAP-KG with BERT for question answering. There was no significant difference observed in performance between question within yes/no and multiclass question groups.}
    \label{tab:qa_results}
\end{table}

\section{Discussion and Future Work}
In this work, we have proposed a novel bespoke knowledge graph extraction method from primary care consultation notes and demonstrated that it outperforms BERT pre-trained for general question answering, while further offering a fully interpretable pipeline, unlike the black-box language models. Interpretability is a valuable property for healthcare since it provides ease of validation when collaborating with domain experts and increases clinician trust. Therefore, RECAP-KG is a promising method for consolidating knowledge from primary care consultation notes. However, at present, our work possesses several limitations that need to be addressed. Firstly, the restricted nature of the data that our model is currently applied to (passages that contain at least one sentence that can be extended to the form 'Patient has ..') does not reflect the full data, resulting in information loss due to sentences that do not fit this form not being processed. This can be addressed by extending the supported sentence structures, as in consultation notes they are typically one of a finite set ('Patient has to..', 'Patient is..', 'Patient's X is Y' etc.). Alternatively, the method can be improved by creating a sentence structure-agnostic pre-processing method. Secondly, we rely on AllenNLP's\cite{Gardner202018AllenNLP:Platform} constituency parser that uses SpaCy\cite{spacy} parse trees - SpaCy relies on a neural network for sentence parsing, which inevitably creates some inconsistencies in creating parse trees (e.g. incorrect part-of-speech) and may thus affect the performance of our model. This can be addressed by adding additional constraints on the constituency parser.

While our model demonstrates strong superiority over BERT for question answering trained on SQuAD\cite{Rajpurkar2016SQuAD:Text}, clinical versions of BERT are readily available. We have therefore fine-tuned BlueBERT (NCBI BERT) - a clinical version of BERT pre-trained on PubMed abstracts and MIMIC-III clinical
notes\cite{Kades2021AdaptingStudy} for question answering. We have trained the model on the SQuAD dataset\cite{Rajpurkar2016SQuAD:Text} for 3 epochs with BertAdam optimiser with a learning rate 2e-05, train and evaluation batch size of 16. We intend to compare our method to this stronger benchmark as part of our future work. Finally, evaluating the performance on multilabel questions, such as 'What symptoms does the patient have?' would provide more context on the text comprehension of RECAP-KG.

The ultimate goal of our research is to produce knowledge graphs for providing more nuanced information about a patient to a CDSS. Knowledge graphs extracted from datasets containing multiple patient entries can be used to infer new information about diseases, e.g. novel coronavirus (COVID-19). Furthermore, automatic clinical decision system generation from knowledge graphs is an avenue we would like to explore. We have already developed a mechanism for translating RECAP-KG knowledge graphs into a symbolic representation. We envision building on this work and applying symbolic learning\cite{Law2018InductiveExamples} to a consolidation of facts extracted from the knowledge graphs to learn new insight about COVID-19 and to automatically generate CDSS support rules.

\bibliographystyle{unsrtnat}
\bibliography{uncap, extra}

\section{Acknowledgements}
This project was supported by UK Research and Innovation. [UKRI Centre for Doctoral Training in AI for Healthcare grant number EP/S023283/1]
\end{document}